\documentclass[11pt]{article}
\usepackage[]{eacl2023}

\usepackage{times}
\usepackage{latexsym}
\usepackage[T1]{fontenc}
\usepackage[utf8]{inputenc}
\usepackage[longtable]{multirow}
\usepackage{longtable}
\usepackage{makecell}
\usepackage{comment}
\usepackage{adjustbox}
\usepackage{enumitem}
\usepackage{amsmath}
\usepackage{tipa}
\usepackage{booktabs}
\usepackage{colortbl}
\usepackage{diagbox}
\usepackage{fixltx2e}
\usepackage{pifont}
\usepackage{amssymb}

\usepackage{arydshln}
\usepackage{float}
\usepackage{afterpage}
\usepackage{capt-of}
\usepackage{tcolorbox}

\usepackage{array}

\newcolumntype{P}[1]{>{\centering\arraybackslash}p{#1}}

\usepackage{makecell}

\usepackage{mathtools}
\usepackage{blkarray, bigstrut} 
\usepackage{microtype}
\usepackage{inconsolata}

\usepackage[normalem]{ulem}

\usepackage{bold-extra}

\usepackage{tikz}
\usetikzlibrary{positioning,arrows.meta,shapes,backgrounds,fit}
\usepackage{listings}
\usepackage{xcolor}

\definecolor{lightbluesparql}{RGB}{225,245,254}
\definecolor{darkbluesparql}{RGB}{1,87,155}
\definecolor{lightpurplesparql}{RGB}{232,234,246}
\definecolor{darkpurplesparql}{RGB}{40,53,147}

\lstdefinestyle{sparql}{
  basicstyle=\ttfamily\footnotesize,
  columns=fullflexible,
  keepspaces=true,
  breaklines=true,
  commentstyle=\color{gray},
  keywordstyle=\color{blue},
  stringstyle=\color{darkpurplesparql},
  showstringspaces=false
}

\newcommand{\nlquery}[1]{\textit{``#1''}}

\title{Conversational Lexicography: Querying Lexicographic Data on \\Knowledge Graphs with SPARQL through Natural Language} 

\author{Kilian Sennrich\\
    Department of Informatics\\
    University of Zurich\\
    \texttt{kilian.sennrich@uzh.ch}
    \And Sina Ahmadi\\
  Department of Computational Linguistics \\
  University of Zurich \\
  \texttt{sina.ahmadi@uzh.ch}}

\begin{document}

\maketitle

\begin{abstract}
Knowledge graphs offer an excellent solution for representing the lexical-semantic structures of lexicographic data. However, working with the SPARQL query language represents a considerable hurdle for many non-expert users who could benefit from the advantages of this technology. This paper addresses the challenge of creating natural language interfaces for lexicographic data retrieval on knowledge graphs such as Wikidata. We develop a multidimensional taxonomy capturing the complexity of Wikidata's lexicographic data ontology module through four dimensions and create a template-based dataset with over 1.2 million mappings from natural language utterances to SPARQL queries. Our experiments with GPT-2 (124M), Phi-1.5 (1.3B), and GPT-3.5-Turbo reveal significant differences in model capabilities. While all models perform well on familiar patterns, only GPT-3.5-Turbo demonstrates meaningful generalization capabilities, suggesting that model size and diverse pre-training are crucial for adaptability in this domain. However, significant challenges remain in achieving robust generalization, handling diverse linguistic data, and developing scalable solutions that can accommodate the full complexity of lexicographic knowledge representation.

\vspace*{-0.2cm}
\begin{center}
\includegraphics[height=0.5cm]{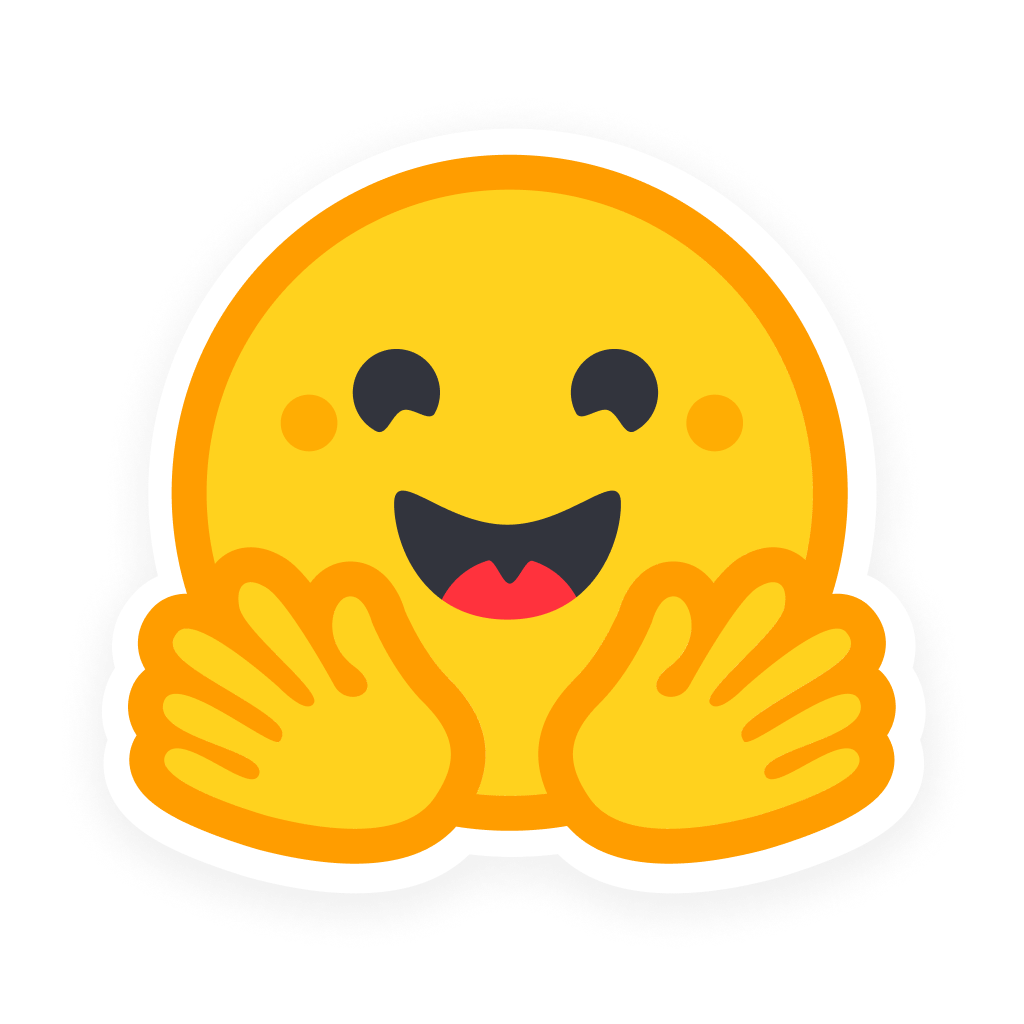}~
\href{https://huggingface.co/datasets/ksennr/lexicographicDataSPARQL}{\textbf{Dataset}} $|$ Models (
\href{https://huggingface.co/ksennr/phi-1_5-finetuned-SPARQLWikidata}{\textbf{Phi-1.5}} $|$ 
\href{https://huggingface.co/ksennr/gpt-2-lexicographic_data_SPARQL}{\textbf{GPT-2}}
)
\end{center}
\end{abstract}

\section{Introduction}

Knowledge Graphs (KGs) have emerged as scalable and interoperable resources for organizing and accessing the vast volumes of data produced in our digital age. Particularly for lexicographic data, as found in dictionaries, KGs offer an ideal structure for capturing the complex relationships between words, meanings, and linguistic patterns due to the highly interrelated nature of this information \citep[p.~14]{ahmadi2022monolingual}. The preservation and accessibility of lexicographic data is crucial for standardizing language understanding, supporting linguistic research, documenting cultural diversity \citep{Gregson2015}, and crucially, increasing interoperability in language technology. Recent advancements in Large Language Models (LLMs) have opened new pathways for creating natural language interfaces to KGs, potentially democratizing access to this structured linguistic knowledge \citep{avila2024experiments}.

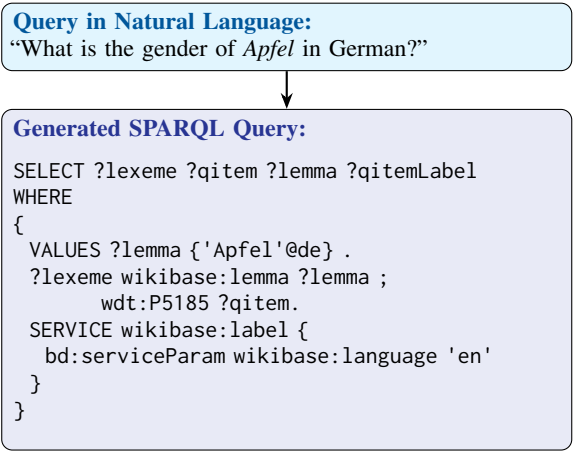
\begin{figure}[t]
  \centering
  \begin{tikzpicture}[
    node distance=1cm,
    box/.style={rectangle, rounded corners, draw, minimum width=\columnwidth-0.2cm, align=left, font=\small},
    title/.style={font=\small\bfseries},
    arrow/.style={-{Stealth[length=6pt]}, thick},
    ]
    
    \node[box, fill=lightbluesparql, text width=\columnwidth-0.5cm] (query) {
      \textbf{\textcolor{darkbluesparql}{Query in Natural Language:}}\\
      ``What is the gender of \textit{Apfel} in German?''
    };

    \draw[arrow] (query.south) -- ++(0,-0.5);
    
    \node[box, fill=lightpurplesparql, text width=\columnwidth-0.5cm, below=0.5cm of query, align=left] (sparql) {
      \textbf{\textcolor{darkpurplesparql}{Generated SPARQL Query:}}\\
      \begin{lstlisting}[style=sparql]
SELECT ?lexeme ?qitem ?lemma ?qitemLabel
WHERE
{
  VALUES ?lemma {'Apfel'@de} .
  ?lexeme wikibase:lemma ?lemma ;
           wdt:P5185 ?qitem.
  SERVICE wikibase:label { 
    bd:serviceParam wikibase:language 'en' 
  }
}\end{lstlisting}
    };
    
  \end{tikzpicture}
  \vspace*{-0.1cm}
  \caption{Conversational lexicography: enabling natural language queries to KGs by automatically generating SPARQL code, eliminating the need for manual query writing}
  \label{fig:conv-lex-example}
\end{figure}

Despite their advantages, KGs remain largely inaccessible to non-technical users due to the specialized knowledge required to query them effectively. Currently, accessing information in KGs requires proficiency in a query language, notably SPARQL, which presents a significant barrier to entry. Users must not only master this technical query language but also understand the specific ontologies and data models that structure each KG \citep{ngonga2013sorry}. Wikidata\footnote{\url{https://www.wikidata.org}}, a prominent open-source KG, employs a collaboratively developed semantic structure that requires detailed knowledge to navigate effectively. This technical complexity limits the broader utility of KGs, particularly for audiences such as language learners, teachers, and other non-technical stakeholders who could benefit from lexicographic data access \citep{warren2020comparison}.

This paper addresses the significant research gap in creating effective natural language interfaces for lexicographic data retrieval on KGs such as Wikidata. To that end, we develop a multidimensional taxonomy that captures the complexity of Wikidata's lexicographic data ontology module, systematically categorizing the diverse information requests that may be queried on the KG. Additionally, we create a template-based dataset that maps natural language utterances to corresponding SPARQL queries, designed to reflect the variety of possible information requests identified in our taxonomy. Finally, we conduct preliminary experiments using transformer-based language models of modest parameter sizes to generate SPARQL queries from natural language inputs, as exemplified in Figure~\ref{fig:conv-lex-example}, evaluating their performance on both seen and unseen utterances to assess the impact of model parameter size and training method. 

\section{Related Work}
\label{sec_related_work}
The translation of natural language queries into SPARQL has received significant attention in recent years, particularly with the advent of LLMs and the increasing importance of KGs. This section provides a brief description of datasets, generation techniques and evaluation methods.

\paragraph{Datasets} The development of specialized datasets has accelerated progress in natural language interfaces to KGs. The Question Answering over Linked Data (QALD) series represents a foundational contribution, with QALD-10 offering the most recent iteration supporting both DBpedia and Wikidata queries \citep{Usbeck2023}. Building on this foundation, the Large-Scale Complex Question Answering Dataset (LC-QuAD 2.0) expands the scope with 30,000 natural language utterances paired with corresponding SPARQL queries \citep{lcquad}. The DBpedia Natural Language Question Answering (DBNQA) dataset stands as one of the most comprehensive resources, containing nearly 900,000 data tuples for training and evaluation \citep{hartmann2018generating}. Addressing the critical need for cross-domain generalization, \citet{kosten2023spider4sparql} introduce Spider4SPARQL with over 10,000 manually crafted SPARQL queries. Experimental evaluations using LLMs demonstrate that Spider4SPARQL presents substantial challenges in achieving high accuracy. 

\paragraph{Generation} Approaches to generating SPARQL queries from natural language have evolved from traditional machine learning to increasingly sophisticated neural architectures. Early work by \citet{DBLP:journals/corr/abs-1806-10478,DBLP:conf/i-semantics/SoruMMPVEN17} establish the foundational \textit{Neural SPARQL Machine} paradigm, comprising a template-based \textit{generator}, a sequence-to-sequence \textit{learner}, and an \textit{interpreter} that translates user inputs into SPARQL. Alternative approaches leverage structural properties of KGs to extract potential RDF triples \cite{DBLP:journals/tkde/Hu0YWZ18,DBLP:journals/jwe/LinL22}, while subsequent advances explore diverse neural architectures, including pre-trained models like BART and T5 \citep{banerjee2022modern}. A persistent challenge is handling incomplete vocabulary, particularly entity identifiers in KGs, e.g., Wikidata's \texttt{Q811486} for `tree', that may not appear during training; researchers have addressed this through Named Entity Disambiguators \citep{LLMsForSparql} and entity masking techniques. For specialized domains, \citet{zou2021chinese} develope a text-to-SPARQL model utilizing a pointer network-based encoder with relation-aware attention mechanisms, while \citet{app14041521} introduce Triplet Structure Enhanced T5, which undergoes a specialized pre-training phase to better handle complex query structures. The emergence of LLMs has further transformed this landscape \citep{perevalov2024towards}. \citet{dabramo-etal-2025-investigating} apply in-context learning using Mixtral (8x7B), Llama-3 (70B), and CodeLlama (70B) to achieve state-of-the-art results, while other approaches demonstrate success through fine-tuning \citep{brei2024leveraging} and one-shot learning \citep{pliukhin2023improving}. \citet{rony2022sgpt} propose SGPT, employing transformer encoders with GPT-2 as the decoder and entity placeholders for post-processing. 

\begin{figure*}
    \centering
    \includegraphics[width=1\linewidth]{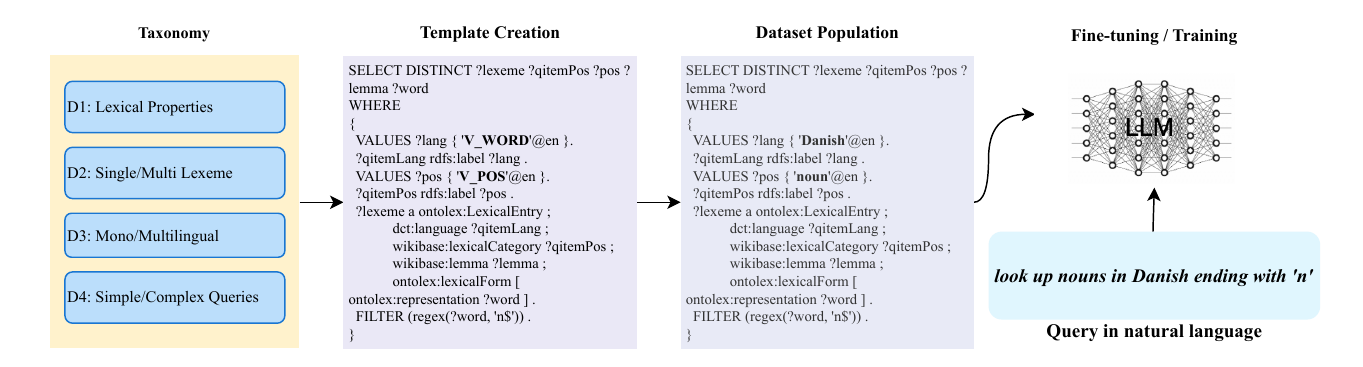}
    \caption{Our approach to creating SPARQL templates based on a four-dimension taxonomy followed by dataset population and model implementation. The ultimate goal is to infer the models by querying in natural language.}
    \label{fig_methodology}
    \vspace*{-0.25cm}
\end{figure*}

\paragraph{Evaluation} The evaluation of natural language to SPARQL systems has traditionally relied on metrics such as accuracy, BLEU~\cite{papineni2002bleu}, F1-score, or a combination of those~\cite{rony2022sgpt}. However, these metrics have limitations, as syntactically different queries can produce identical results. \citep{evaluation2013} propose evaluation frameworks that combine syntactic metrics with semantic correctness assessments to capture the practical utility of generated queries. Recent work suggests moving beyond simple comparison with gold standards toward functional correctness testing \citep{Chen2021EvaluatingLL}, similar to general code generation evaluation approaches.

As such, several research gaps persist in this domain. First, existing datasets predominantly focus on factual knowledge, leaving lexicographical queries underexplored. Second, the optimal approach to handling incomplete vocabulary and generalization remains an open question. Finally, while LLMs show promise for SPARQL generation, their potential specifically for lexicographic data queries remains uncertain. 

\section{Methodology}
We develop a systematic methodology to map natural language queries to SPARQL for lexicographic data in Wikidata, illustrated in Figure~\ref{fig_methodology}. This relies on a taxonomy to generate query templates which are then populated with data instances to create a comprehensive dataset. The dataset is subsequently used to train and fine-tune LLMs for the SPARQL generation task. We provide background information about Wikidata in Appendix~\ref{appendix_background}.

\subsection{Taxonomy for the Lexicographic Data}
\label{subsec_taxonomy}
To systematically approach template creation for lexicographic data, we develop a taxonomy that defines the relevant aspects of translating natural language to SPARQL queries in Wikidata's lexicographic domain. Our taxonomy is based on three criteria:

\begin{itemize}[noitemsep]
\item[] \textbf{Criterion 1}: It should encompass the full range of SPARQL syntax constructs and operators\label{crit1}
\item[] \textbf{Criterion 2}: It should cover the variety of use cases for lexicographic data \label{crit2}
\item[] \textbf{Criterion 3}: It should be particularly detailed in frequently queried areas \label{crit3}
\end{itemize}

These criteria guided the identification of four feature dimensions (D) that capture the heterogeneity of lexicographic queries:

\paragraph{D1: Lexical Properties} This dimension addresses Criterion 2 by covering the range of lexicographic properties in Wikidata. These properties serve as fundamental building blocks for SPARQL queries using the lexicographic data ontology module. We classify these properties into the following seven categories, summarized in Table~\ref{tab:propertyClassification} in the appendix:

\begin{itemize}[noitemsep,nolistsep]
    \item \textit{Linguistic Properties}: Grammatical and morphological features, e.g., grammatical gender, conjugation class
    \item \textit{Historical References}: Temporal aspects of lexemes, e.g., first attestation
    \item \textit{Syntactic Functions}: Roles of lexemes within sentences, e.g., auxiliary verb, examples
    \item \textit{Semantic Relations}: Meaning relationships between lexemes, e.g., synonyms, antonyms
    \item \textit{Orthographic and Phonetic Features}: Written and spoken forms, e.g., IPA transcription
    \item \textit{Translation and Lexical Variety}: Cross-linguistic information and variants, e.g., borrowed forms, regional variants
    \item \textit{Stylistic Attributes}: Context-dependent characteristics, e.g., language register, tone
\end{itemize}

\paragraph{D2: Single vs. Multi Lexeme Output} This dimension focuses on whether the natural language query targets a single lexeme or multiple lexemes. This classification is based on the semantics of the utterance rather than the actual number of lexemes in the output. For example, the question \nlquery{What is the grammatical gender of the French word `livre'?} is classified as Single-Lexeme Output despite potentially returning multiple homograph lexemes  (masculine `\textit{livre}' meaning `book' and feminine `\textit{livre}' meaning `pound' as unit of weight). This dimension is particularly important for addressing Criterion 1, as certain SPARQL keywords and structures are associated with either Single- or Multi-Lexeme queries. Conversely, some utterances inherently imply a Multi-Lexeme Output. An example is the utterance \nlquery{Create a French-German-Basque lexicon}.

\paragraph{D3: Mono- vs. Multilinguality} This dimension distinguishes between queries that involve one language versus those that involve multiple languages. Classification is based on the languages of all lexemes that would appear in the output if all variables were included. For instance, the query \nlquery{What is the French word for `fish'?}, is classified as multilingual because lexemes from multiple languages appear in the result. This dimension addresses Criterion 3. 

\paragraph{D4: Simple vs. Complex Queries} This dimension analyzes query complexity based on the number of lexical properties involved. While ``complex'' in literature often refers to queries requiring multiple reasoning steps \cite{Multi_step_reasoning}, we define simple queries as those containing only one lexical property, e.g., \nlquery{From what word is the French word `\textit{cigare}' derived?}, and complex queries as those containing multiple properties. This definition better suits lexicographic data, where users target properties of a single lemma rather than performing multi-step reasoning.

\subsection{Implementation}
We implement two distinct approaches to fine-tune and train models for natural language to SPARQL:

\begin{itemize}
    \item First, we fine-tune a pre-trained Phi-1.5 model \citep{li2023textbooks} using the Low-Rank Adaptation (LoRA) framework. Phi-1.5 is a small language model with 1.3B parameters that demonstrates strong capabilities in both natural language and code generation. For fine-tuning, we use the following hyperparameters: learning rate of 0.0002, train batch size of 4, Adam optimizer, cosine learning rate scheduler, and mixed precision training. Following \citet{schimanski2024towards}, we limited training to a single epoch to avoid overfitting. The LoRA approach allowed us to fine-tune 0.44\% of the model's parameters.
    \item Second, we train a GPT-2 architecture with 124M parameters \citep{GPT-2} from scratch using the Hugging Face library. For this model, we use a learning rate of 5e-05, train batch size of 16, Adam optimizer, linear learning rate scheduler, and trained for three epochs.
\end{itemize}

Both models are trained on data formatted by concatenating natural language utterances prefixed with ``\texttt{question:}'', and corresponding SPARQL queries prefixed with ``\texttt{answer: <code>}'' and ``\texttt{suffixed with "</code>"}''. This format simplifies the parsing of SPARQL code from the output. The training utilized Phi-1.5's tokenizer, which extends GPT-2's BPE vocabulary with special tokens for code representation. We employ two NVIDIA GeForce RTX 3090 GPUs with CUDA 12.4 for training.

\subsection{Evaluation}
\label{subsection_evaluation}

Inspired by \citet{evaluation2013}, we deploy an evaluation framework structured around the following four key principles: 

\begin{enumerate}[label=\Alph*.,noitemsep,nolistsep,leftmargin=*]
    \item \textbf{Automatic evaluation} of the text-to-SPARQL model rather than manual; 
    \item \textbf{Functionality} prioritizing functional correctness over exact match, i.e., character-by-character comparison of the generated SPARQL query with a gold standard reference query. In our evaluation setup, we use \citet{Chen2021EvaluatingLL}'s $pass@k$ metric which generates $k$ responses for a given prompt containing few-shot examples. Each of the generated responses is then run against the KG.\footnote{Wikidata Query Service: https://query.wikidata.org} If the triples retrieved by the generated query match or include the expected answer triples from the gold standard query, the generated response is deemed correct. The $pass@k$ metric is then calculated as the ratio of all the correctly generated responses ($k_{\text{correct}}$) within the $k$ trials and all generated responses:
    \begin{equation}
        pass@k = \frac{k_{correct}}{k} 
    \end{equation}
    
    \item \textbf{Granularity} employing unit test-like checks to evaluate specific aspects of the generated SPARQL queries, including syntax correctness and appropriate variable usage rather than just overall correctness. As such, we define a granularity ratio to assess the fine-grained quality of generated queries as follows:
    \begin{equation}
        R_{\text{granularity}} = \frac{c_{\text{pass}}}{c_{\text{all}}} 
    \end{equation}
    where $c_{\text{pass}}$ is the number of passed checks and $c_{\text{all}}$ is the total number of checks performed. A list of the tests is provided in Appendix~\ref{app_sec_evaluation}.
    \item \textbf{Generalization} assessing the model's ability to generalize by altering input questions to trigger different query types. To do so, we transform a training question like \nlquery{What is the gender of `Apfel' in German?} (requiring a \texttt{SELECT} query) into a test question like \nlquery{Is the gender of `Apfel' in German feminine?} (requiring an \texttt{ASK} query), testing whether the model can adapt to this structural change.
\end{enumerate}

Finally, for string-based matching, we report performance using BLEU as implemented in SacreBLEU~\cite{post-2018-call}.\footnote{\tiny{\texttt{nrefs:1|case:mixed|eff:no|tok:13a|smooth:exp|version:2.4.2}}}

\section{Dataset}
To develop a comprehensive dataset mapping natural language utterances to SPARQL queries targeting lexicographic data in Wikidata, we adopt a template-based approach similar to \citet{DBLP:conf/i-semantics/SoruMMPVEN17} based on the taxonomies defined in Section \ref{subsec_taxonomy}. Each data point in our templates consists of three elements: 

\begin{enumerate}[noitemsep]
    \item \textbf{utterance}: natural language input reflecting a user's question;
    \item \textbf{template\_name}: identifier for the template in SPARQL containing tags that are later populated with actual words; 
    \item \textbf{query}: the populated SPARQL template aligned with the utterance. 
\end{enumerate}

All utterances are in English, though they may reference terms in other languages, e.g., \nlquery{What is the grammatical gender of `livre' in French?}. The following is an instance in our populated dataset:

\begin{tcolorbox}[
  colback=lightgray!10, 
  colframe=gray!50, 
  left=5pt,
  right=5pt,
  top=5pt,
  bottom=5pt,
  boxsep=2pt,
  fontupper=\small
]
\textbf{utterance}: where does the word color come from?

\textbf{template\_name}: q20

\textbf{query}: 
\begin{verbatim}
SELECT ?etonymLexeme ?qitemLanguageOfOrigin 
       ?etonym ?qitemLanguageOfOriginLabel
WHERE {
  VALUES ?lemma {'color'@en} .
  ?lexeme wikibase:lemma ?lemma ;
          wdt:P5191 ?etonymLexeme.
  ?etonymLexeme dct:language ?qitemOrigin;
                wikibase:lemma ?etonym .
  SERVICE wikibase:label { 
    bd:serviceParam wikibase:language 'en' 
  }
}
\end{verbatim}
\end{tcolorbox}

To address the limited diversity inherent in template-based approaches, we decouple semantics from syntax by generating multiple variations of utterance templates while preserving their meaning. This is accomplished by using GPT-4 to generate alternative phrasings with random selection during template population with an example provided in Appendix~\ref{app_sub_prompt}. 

\subsection{Template Sources}

Our dataset comprises five specialized modules following different paradigms:

\paragraph{Google Templates} Following \citet{hazoom2021text}, who advocate deriving data from naturalistic environments, we extract questions related to lexicographic data from Google's Natural Questions dataset. We identify relevant lexicographic terms and extract 3,296 user questions containing these terms. To do so, we cluster questions using $k$-means and \texttt{FlagEmbeddings} embedding model~\cite{bge_m3}\footnote{\texttt{BAAI's BGE-Large} variant}. We then manually review clusters to identify 639 genuinely relevant questions. The selected questions yield 21 unique SPARQL templates that closely align with typical user questions (see Appendix~\ref{tab:clusterUtterances} for sample cluster). Analysis of the Natural Questions dataset showed 35\% multilingual vs. 65\% monolingual and 52\% complex vs. 48\% simple queries, informing our template distribution to meet Criterion 3.

\paragraph{Property Templates} To enable efficient Wikidata usage through natural language interfaces, we also create templates covering properties specific to the WikibaseLexemes extension. We manually select 36 relevant properties from lexicographical properties, categorizing them based on their domain (lexeme, sense, or form) and range data type (string, Q-item, etc.). This dual classification resulted in nine archetypal SPARQL templates, which are further adapted to handle multi-lexeme outputs and ASK statements.

\paragraph{Multi-Property Templates} These templates address queries requiring multiple pieces of information for a given lexeme. All multi-property queries derive from a single adjustable base template modified to handle both single-result and multiple-result queries. The templates use the \texttt{OPTIONAL} keyword to handle cases where properties are unavailable for certain lexemes. Properties are randomly selected from a pool of 211 options (not restricted to WikibaseLexemes) to prevent overfitting. Two versions of utterance templates were used: single-lexeme and multi-lexeme.

\begin{table*}[t]
\centering
\begin{tabular}{ccccccccc} 
\toprule
\multirow{2}{*}{Model} & \multirow{2}{*}{Parameter} & \multicolumn{3}{c}{Non-Generalization} & \multicolumn{3}{c}{Generalization} \\ \cmidrule(lr){3-5} \cmidrule(lr){6-8}
  &  & $pass@k$$\uparrow$ & $R_{\text{granularity}}$$\uparrow$ & BLEU$\uparrow$ & $pass@k$$\uparrow$ & $R_{\text{granularity}}$$\uparrow$ & BLEU$\uparrow$\\ \midrule
    Phi 1.5 & $k$=1 & 0.86 & 0.84 & 92.1 & 0 & 0.7 & 54.4\\
  GPT-2 & $k$=1 & 0.90 & 0.84 & 94.4 & 0 & 0.41 & 0.3\\ \hline
 \multirow{2}{*}{GPT-3.5 Turbo} & $k$=1 & 0.87 & 0.94 &  99.2& 0.41 & 0.81 & 72.7\\
  & $k$=3 & 0.89 & 0.95 &  99.6& 0.57 & 0.84 & 67.0\\ 
\bottomrule
\end{tabular}
\caption{Performance of few-shot fine-tuned GPT-3.5 Turbo in comparison to our trained and fine-tuned models using $pass@k$ \([0, 1]\)  for functionality, $R_{\text{granularity}}$ \([0, 1]\) for granularity and BLEU \([0, 100]\). Although GPT-3.5 Turbo as the baseline performs better than our models, our trained GPT-2 model achieves a higher $pass@k$ despite having significantly less parameters. Due to computational costs, $k=3$ could not be included for Phi 1.5 and GPT-2.}
\label{tab_experiments_results}
\end{table*}

\paragraph{Language-Independent Templates} These templates function without specifying the lexeme's language, enabling cross-language lookups. They use string matching (\texttt{FILTER(STR(?lemma) = "word")}) rather than language-specific \texttt{VALUES} clauses, trading computational efficiency for flexibility. Since these queries can return numerous lexemes, we introduced templates restricting output based on lexical category and grammatical features. This resulted in eight templates covering both language-dependent and language-independent queries.

\paragraph{Rule-Based Templates} This paradigm incorporates existing work in lexicographic data querying. We adapted seven templates from SPARQLify\footnote{\url{https://sinaahmadi.github.io/SPARQLify}}, a simple form-based query generator. These templates cover advanced use cases employing multiple properties and SPARQL functions as in \texttt{regex()}) not represented in other paradigms, such as \nlquery{Find at most 50 longest words in \{language\}} and \nlquery{List at most 50 onomatopoeia in \{language\}}.

\subsection{Dataset Population}

We populate templates by replacing tags with actual lemmas from Wikidata, ensuring that lexemes had relevant properties whenever possible. The data used represents a snapshot from April-May 2024, constrained by Wikidata's query limits (30,000 data points maximum, one-minute computation time). A custom Python program replaced template tags with corresponding population data.

\begin{figure}[t]
  \centering
  {\includegraphics[width=1\columnwidth]{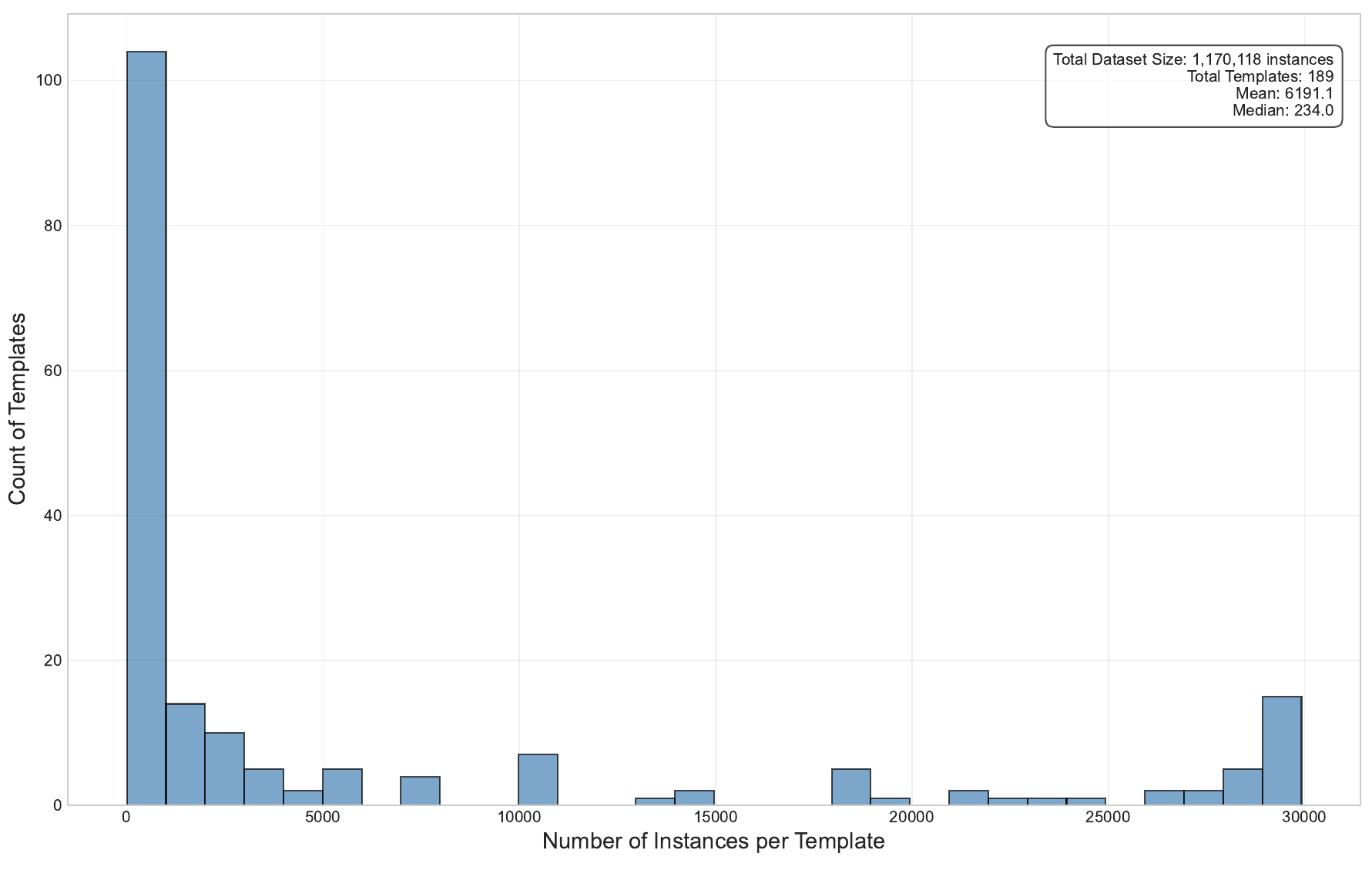}}
  \caption{Distribution of the number of populated data tuples per template}
  \label{fig:distDataTuples}
\end{figure}

\subsection{Dataset Statistics}

Our dataset contains 1,270,113 data tuples derived from 189 templates with an average of 6,191 instances per template. Templates populated between 1 (for \textit{limit\_t9\_P2859} and \textit{order\_t9\_P2859}) and 29,922 (for \textit{ask\_t9\_P7243} and \textit{t9\_P7243}) data tuples each. Approximately half of the templates populated over 1,000 data tuples. The distribution of the number of populated tuples per template is illustrated in Figure \ref{fig:distDataTuples}. Following \citet{DBLP:conf/i-semantics/SoruMMPVEN17}, we define the train-test split such that the evaluation dataset contains at most 10\% of data points per template, with a maximum of 20 data points. This ensures a balanced evaluation set while maintaining a substantial training set. From our dataset, we include at least one instance of each template in the test set to ensure comprehensive evaluation.

\section{Experiments and Results}

In order to evaluate the effectiveness of various language models in generating SPARQL queries for lexicographic data on Wikidata, we conduct experiments with three strategically selected models: GPT-3.5-Turbo as a \textbf{baseline}, and our \textbf{fine-tuned} Phi-1.5 and \textbf{trained} GPT-2 models. When evaluated in a zero-shot setting without fine-tuning or training, both Phi-1.5 and GPT-2 failed completely, scoring 0 across all metrics, demonstrating that task-specific adaptation is essential for SPARQL generation with these models.

Our selection of models prioritizes those with modest parameter counts (1.3B for Phi-1.5 and 124M for GPT-2) to demonstrate if effective SPARQL generation can be achieved without requiring computationally expensive models, making deployment more accessible for resource-constrained environments. Additionally, these models represent different training approaches--GPT-3.5-Turbo as a commercial API-based model, Phi-1.5 as a recent code-capable model amenable to parameter-efficient fine-tuning, and GPT-2 as a fully trainable smaller model--providing a diverse evaluation spectrum. For each model, we assess performance using the evaluation framework described in Section~\ref{subsection_evaluation}. The results are summarized in Table \ref{tab_experiments_results}.

\subsection{GPT-3.5-Turbo}

We evaluate GPT-3.5-Turbo to establish a baseline against which our custom-trained models can be compared. Despite its extensive parameter count, this model performs poorly when directly asked to generate lexicographic SPARQL queries. We leverage GPT-3.5-Turbo's strong few-shot learning capabilities by employing prompt engineering, sampling two random utterances and corresponding SPARQL queries from the training dataset for each template to create the prompt, with an example in Appendix \ref{app_baseline_prompt}.

In the evaluation without generalization, GPT-3.5-Turbo achieves a $pass@1$ score of 0.87 and $R_{\text{granularity}}$ of 0.94. When allowed to generate multiple responses ($k=3$), performance improves to 0.89 and 0.95 respectively. For the evaluation with generalization, performance drops to a $pass@1$ score of 0.41 and $R_{\text{granularity}}$ of 0.81, improving to 0.57 and 0.84 with $k=3$, highlighting the challenge of adapting to novel query structures. The same pattern is seen in BLEU scores, except in generalization where the BLEU score with $k=3$ (67.0) is lower than $k=1$ (72.7). This counter-intuitive result can be explained by the model's tendency to explore more diverse, but potentially less syntactically aligned, query structures when generating multiple responses. While this diversity improves functional correctness (as measured by $pass@k$), it reduces strict textual similarity to reference queries. 

\subsection{Phi 1.5}

We evaluate Phi-1.5 fine-tuned on our dataset with $k=1$ only, a decision driven by significant computational demands—the evaluation without generalization alone requires 23 hours to complete. The model achieves a $pass@1$ score of 0.86 and $R_{\text{granularity}}$ of 0.84 in non-generalization scenario.

Our analysis indicates that Phi-1.5 does not attempt to generalize beyond specific SPARQL structures from fine-tuning. While information from utterances is correctly mapped to appropriate positions in the code, the query structure remains closely aligned with training examples. In the generalization scenario, the model struggles significantly with a $R_{\text{granularity}}$ of 0.7, indicating that many generated queries fail to meet basic correctness criteria. 

\subsection{GPT-2}

We evaluate GPT-2 trained from scratch on our dataset, representing a model unexposed to any data except our training examples. Similar to Phi-1.5, we compute results with $k=1$ only due to computational constraints. In the evaluation without generalization, GPT-2 achieves the highest $pass@1$ score among all models at 0.90, with a $R_{\text{granularity}}$ of 0.84. In the generalization scenario, however, GPT-2's performance deteriorates substantially, with a $R_{\text{granularity}}$ of only 0.41 and BLEU score of 0.3, the lowest among all models. This suggests a high degree of memorization rather than a deeper understanding of the relationship between natural language and SPARQL. The model's strong performance in familiar scenarios coupled with poor generalization indicates effective pattern learning but limited transfer capability.

\subsection{Qualitative Analysis}
Our qualitative analysis reveals distinct patterns across models. Phi-1.5 demonstrates limited semantic understanding, surprising knowledge of less-resourced language tags, and accurate syntactic mapping, but struggles with generalization, often generating syntactically correct but semantically nonsensical SPARQL code. GPT-2 exhibits similar semantic limitations (interpreting ``lengthy words'' as words with specific prefixes) and contextual failures, but handles special characters well; in generalization, it produces random word sequences and incomplete syntax. GPT-3.5-Turbo occasionally uses incorrect language tags and struggles with special characters, but shows better understanding of complex utterances and develops creative adaptation strategies like nesting \texttt{SELECT} statements within \texttt{ASK} blocks. Overall, few-shot GPT-3.5-Turbo achieves superior performance across most metrics, though trained GPT-2 excels in $pass@1$ for familiar queries despite having significantly fewer parameters. These findings suggest that while smaller models can be effectively trained for domain-specific SPARQL generation within familiar patterns, robust generalization to novel query structures may require larger models with diverse pre-training or more sophisticated fine-tuning approaches.

\section{Conclusion and Discussion}

This paper addresses the challenge of creating natural language interfaces for lexicographic data in KGs. We develop a multidimensional taxonomy capturing the complexity of Wikidata's lexicographic data representation based on which we create a template-based dataset with over 1.2 million mappings from natural language utterances to SPARQL queries. Our experiments with GPT-2, Phi-1.5, and GPT-3.5-Turbo reveal significant differences in model capabilities. While all models perform well on familiar query patterns ($pass@1$ scores ranging from 0.86 to 0.90), only GPT-3.5-Turbo demonstrates meaningful generalization capabilities ($pass@3$ of 0.57 in the generalization scenario). This suggests that model size and diverse pre-training are crucial for adaptability in this domain. This work is timely and important as KGs continue to grow in complexity, creating an urgent need for accessible interfaces.

\paragraph{Limitations and Future Work}
While our experiments demonstrate promising results with models of modest size, future work should explore more recent reasoning-focused models such as DeepSeek, QwQ, and Llama-3, which may offer improved performance for complex SPARQL generation tasks. Additionally, scaling experiments with larger model variants could help establish the relationship between model size and SPARQL generation capabilities, potentially identifying optimal efficiency-performance tradeoffs for this specific task. As such, future work should focus on improving model generalization through more diverse training data, expanding this approach to other KGs, particularly Dbnary~\cite{DBLP:conf/lrec/Serasset12}, and conducting user studies to evaluate practical utility for different stakeholder groups in lexicography and linguistics.

\section*{Acknowledgments}
This work was supported by the Swiss National Science Foundation (MUTAMUR; no. 213976). The authors are grateful to the constructive feedback of the anonymous reviewers.

\vspace{-0.5cm}
\bibliography{references}
\bibliographystyle{acl_natbib}

\appendix
\counterwithin{figure}{section}
\counterwithin{table}{section}
\label{sec:appendix}

\section{Prompts}
\label{app_sub_prompt}
\begin{tcolorbox}[colback=lightgray!10, colframe=gray!50, title=ChatGPT prompt for utterance diversification,  fonttitle=\small,   fontupper=\small]
\textbf{Input:} Rewrite the following sentence 10 times in a way someone may ask it on Google. Keep it short: Look up \{\textit{word}\} in any language in \{\textit{form\_word}\} form

\textbf{Output:}
\begin{itemize}[noitemsep]
    \item Translate \{\textit{word}\} to any language in \{\textit{form\_word}\}?
    \item How to say \{\textit{word}\} in \{\textit{form\_word}\} across languages?
    \item Find \{\textit{word}\} in any language's \{\textit{form\_word}\}?
    \item Discover \{\textit{word}\} in different languages, \{\textit{form\_word}\}?
    \item Search for \{\textit{word}\} in \{\textit{form\_word}\} form in all languages?
    \item Get \{\textit{word}\} translations in \{\textit{form\_word}\} in any language?
    \item Lookup \{\textit{word}\} in various languages, \{\textit{form\_word}\}?
    \item How to express \{\textit{word}\} in \{\textit{form\_word}\} across languages?
    \item Find \{\textit{word}\} equivalents in \{\textit{form\_word}\} across languages?
    \item Translate \{\textit{word}\} into \{\textit{form\_word}\} in any language?
\end{itemize}
\end{tcolorbox}

\vspace{1cm}

\begin{tcolorbox}[colback=lightgray!10, colframe=gray!50, title=Few-Shot Prompt Example for GPT-3.5 Turbo, fonttitle=\small,fontupper=\small]
\label{app_baseline_prompt}
\begin{verbatim}
Utterance 1:
Apfel gender in German
SPARQL 1:
SELECT ?lexeme ?qitem ?lemma ?qitemLabel
WHERE
{
  VALUES ?lemma {'Apfel'@de} .
  ?lexeme wikibase:lemma ?lemma ;
          wdt:P5185 ?qitem.
  SERVICE wikibase:label { 
    bd:serviceParam wikibase:language 'en' 
  }
}

Utterance 2:
medailon gender Czech
SPARQL 2:
SELECT ?lexeme ?qitem ?lemma ?qitemLabel
WHERE
{
  VALUES ?lemma {'medailon'@cs} .
  ?lexeme wikibase:lemma ?lemma ;
          wdt:P5185 ?qitem.
  SERVICE wikibase:label { 
    bd:serviceParam wikibase:language 'en' 
  }
}

Utterance:
What is Probekörpers gender in German?
\end{verbatim}
\end{tcolorbox}

\newpage

\section{Lexicographical Data on Wikidata}
\label{appendix_background}
This section provides essential background on lexicographic data and its representation on Wikidata.

\subsection{Lexicographic Data}
Lexicography is the field concerned with dictionaries and reference works. Lexicographic data encompasses all information contained within dictionaries or reference works, which may range from traditional print dictionaries to digital databases and KGs. The ontology for lexicographic data on the Semantic Web is primarily supported by OntoLex-Lemon \citep{mccrae2017ontolexlemon}, which is based on the Lexicon Model for Ontologies (lemon). This model relies on LexInfo \citep{cimiano2011lexinfo}, LMF \citep{francopoulo2006lexical}, and LIR \citep{montiel2008modelling}. The OntoLex lexicography module, known as \textit{lexicog} \citep{bosque2017towards}, provides key concepts like \textit{LexicalEntry} and \textit{LexicalSense} that were influential in Wikidata's development. Wikidata has expanded beyond representing concepts to include structured descriptions of words through lexemes, forms, and senses. The lexicographic data module follows the \textit{Wikibase} data model, extended with the WikibaseLexemes ontology module that introduces the data types \textit{Lexemes}, \textit{Forms}, and \textit{Senses}.

\paragraph{Lexemes} A lexeme is a fundamental vocabulary unit that can take various forms including simple words, complex words, phrasal words, and multi-word expressions. In Wikidata, lexemes have:

\begin{itemize}[noitemsep,nolistsep]
    \item Unique IDs starting with `L', e.g., \texttt{L870817} for `\textit{Steilkurve}' in German
    \item Lemmas providing human-readable representations, e.g., `book'
    \item Language specification using Q-items, e.g., \texttt{Q1860} for English
    \item Lexical category indicated by Q-items, e.g., \texttt{Q34698} for adjective
    \item Statements describing properties not specific to forms or senses
    \item Forms for each combination of grammatical features
    \item Senses describing different meanings
\end{itemize}

\paragraph{Lemmas} A lemma serves as a location pointer for information within a reference work. In Wikidata, lemmas are implemented as \texttt{MultilingualTextValues}\footnote{\url{https://www.mediawiki.org/wiki/Wikibase/DataModel\#MultilingualTextValues}} to accommodate languages with active diagraphia such as Serbian which uses both Cyrillic and Latin alphabets. The canonical form of the lexeme, typically the infinitive form of verbs, is used as the lemma. For example, the lemma for the English noun `\textit{color}' would include both `\textit{colour}' for British English and `\textit{color}' for American English. Further, lemmas are not unique, and the combination of lemma, language, and lexical category is not unique either. For instance, there are two German nouns with the lemma `\textit{See}' that differ only in gender, with `\textit{der See}' meaning `\textit{the lake}' and `\textit{die See}' meaning `\textit{the sea}'. These two meanings cannot be understood as a single lexeme, as they have different forms based on their gender. In RDF, Wikidata lexemes are represented as \texttt{ontolex:LexicalEntry}, connected to their senses with the \texttt{ontolex:sense} property and to their forms with the \texttt{ontolex:lexicalForm} property. Each lexeme has an associated lemma (\texttt{wikibase:lemma}) and language (\texttt{dct:language}).

\paragraph{Senses}
A sense represents one of the multiple meanings a word can have, arising from polysemy or homonymy. In Wikidata, senses are attributed to lexemes and identified by unique IDs (lexeme ID + \texttt{-S} + decimal number as in \texttt{L16168-S1} for the act of booking in the ``book'' lexeme \texttt{L16168}). Each sense typically includes a gloss providing a natural language definition and may have statements describing relationships with other senses and items (synonyms, antonyms, etc.).

\paragraph{Forms}
A form refers to the specific manifestation of a lexeme in a grammatical context. In Wikidata, forms have unique identifiers (lexeme ID + \texttt{-F} + decimal number as in \texttt{L16168-F1} for the simple past of `book') and are characterized by grammatical features and statements providing information about usage, pronunciation, etc.

\paragraph{Properties}
Properties model relationships between subjects and objects in KGs. In Wikidata, properties describe the data value of a statement and have labels, descriptions, and aliases in multiple languages. Each property has a specific data type and a unique identifier with a \texttt{P} prefix. Lexicographic properties are a subset used with the WikibaseLexeme data model.

\newpage
\section{Evaluation}
\label{app_sec_evaluation}

\begin{table}[!hbt]
\setlength{\tabcolsep}{14pt}
\resizebox{\columnwidth}{!}{%
\begin{tabular}{@{}p{1.5cm}p{6cm}@{}}
\toprule
\multicolumn{1}{c}{\textbf{Category}} & \multicolumn{1}{c}{\textbf{Property}} \\ \midrule
\multirow{11}{*}{\centering\begin{tabular}[c]{@{}c@{}}Linguistic\\Properties\end{tabular}} & - grammatical gender (P5185)\\
 & - conjugation class (P5186)\\
 & - word stem (P5187)\\
 & - derived from lexeme (P5191)\\
 & - combines lexemes (P5238)\\
 & - homograph lexeme (P5402)\\
 & - valency (P5526)\\
 & - requires grammatical feature (P5713)\\
 & - paradigm class (P5911)\\
 & - grammatical aspect (P7486)\\
 & - predicate for (P9970)\\ \midrule
\multirow{2}{*}{\centering\begin{tabular}[c]{@{}c@{}}Historical\\References\end{tabular}} & - attested in (P5323)\\
 & - first attested from (P6684)\\ \midrule
\multirow{6}{*}{\centering\begin{tabular}[c]{@{}c@{}}Syntactic\\Functions\end{tabular}} & - auxiliary verb (P5401)\\
 & - classifier (P5978)\\
 & - location of sense usage (P6084)\\
 & - usage example (P5831)\\
 & - creates lexeme type (P5923)\\
 & - false friend (P5976)\\ \midrule
\multirow{5}{*}{\centering\begin{tabular}[c]{@{}c@{}}Semantic\\Relations\end{tabular}} & - synonym (P5973)\\
 & - antonym (P5974)\\
 & - troponym of (P5975)\\
 & - said to be the same as lexeme (P11577)\\
 & - pertainym of (P8471)\\ \midrule
\multirow{5}{*}{\centering\begin{tabular}[c]{@{}c@{}}Orthographic /\\Phonetic\\Features\end{tabular}} & - Han character in this lexeme (P5425)\\
 & - IPA transcription (P898)\\
 & - X-SAMPA code (P2859)\\
 & - Slavistic Phonetic (P5276)\\
 & - pronunciation (P7243)\\ \midrule
\multirow{2}{*}{\centering\begin{tabular}[c]{@{}c@{}}Translation\end{tabular}} & - translation (P5972)\\
 & - variety of lexeme, form or sense (P7481)\\ \midrule
\multirow{3}{*}{\centering\begin{tabular}[c]{@{}c@{}}Stylistic and\\Phonological\\Attributes\end{tabular}} & - language style (P6191)\\
 & - collective noun for animals (P6571)\\
 & - tone or pitch accent class (P5426)\\ \bottomrule
\end{tabular}
}
\caption{A taxonomic classification of Wikidata Lexicographic Properties organized by categories}
\label{tab:propertyClassification}
\end{table}

\newpage

For the granularity test, the following checks are performed:
\begin{itemize}[noitemsep]
    \item The response must start with either \texttt{SELECT} or \texttt{ASK}
    \item If it starts with \texttt{SELECT}, there must be at least one variable starting with ? before the \texttt{WHERE} clause
    \item If it starts with \texttt{ASK}, there must be a \texttt{WHERE} clause following directly after
    \item Every \{ must have a corresponding \}
    \item The response must not contain the keyword \texttt{VALUES}
    \item The response must contain at least one of the following variables: \textit{?lexeme, ?lemma, ?form, ?sense, ?qitem, ?qitemlabel}
    \item The response must not contain any Q-items that are not in the known Q-items
\end{itemize}

\begin{table}[h]
\small
\setlength{\tabcolsep}{2pt}
\resizebox{\columnwidth}{!}{%
\begin{tabular}{@{}ll@{}}
\toprule
\multicolumn{1}{c}{\textbf{Index}} & \multicolumn{1}{c}{\textbf{Utterance}} \\ \midrule
1     & what is the definition of low birth weight              \\
2     & what does the prefix re mean in medical terminology     \\
3     & what does e/m stand for in medical terms                \\
4     & what does ncd stand for in medical terms                \\
5     & what does acs stand for in medical terms                \\
6     & in military terms what does gi stand for                \\
7     & what does pvc stand for in medical terms                \\
8     & what does mi stand for in medical terms                 \\
9     & what is a pa c in medical terms                         \\
10    & what does la stand for in medical terms                 \\
11    & what does ts stand for in medical terms                 \\
12    & how do you write twice a day in medical terms           \\
13    & what does dc stand for in medical terms                 \\
14    & what does ta stand for in medical terms                 \\
15    & what does ibm stand for in medical terms                \\
16    & what is the definition of an asthma attack              \\
17    & what is the full meaning of cpr in first aid            \\
18    & what is the meaning of rx in medical line               \\
19    & meaning of od and bd in medical term                    \\
20    & medical term meaning condition of stones in the ureters \\ \bottomrule
\end{tabular}
}
\caption{Utterances potentially targeting lexicographic information in one of the clusters of the Google Templates. This cluster is dominated by utterances  about medical abbreviations. However, the presence of an utterance discussing military abbreviations (index 6), suggests that the clustering considers not only the topic of the utterance, but also its lexicographical category.}
\label{tab:clusterUtterances}
\end{table}

\end{document}